\documentclass[letterpaper, 10 pt, conference, ngerman]{ieeeconf}  
\pdfoutput=1

\IEEEoverridecommandlockouts

\usepackage[pdftex]{graphicx}
\usepackage{epstopdf}
\usepackage{color}
\usepackage[caption=false,font=footnotesize]{subfig}
\usepackage{amsmath} 
\usepackage{amssymb}
\usepackage{algpseudocode}
\usepackage{lipsum}
\usepackage{multirow}
\usepackage{tabularx}
\usepackage{algorithmicx}
\usepackage{hyperref}
\usepackage{breakurl}
\usepackage{cite}
\usepackage{setspace}

\usepackage{textcomp}
\usepackage[T1]{fontenc}

\newcommand{\norm}[1]{\left\Vert#1\right\Vert}

\newcommand{\bbm}[1]{\begin{bmatrix} #1 \end{bmatrix}}

\newcommand{\bframe}[1]{ \hspace{0pt}^{\mathcal{B}}#1 }

\newcommand{\ebz}{\bframe{{\mathbf{e}}_z}}

\newcommand{\feg}{\mathbf{f}^e}
\newcommand{\taueg}{\boldsymbol\tau^e}
\newcommand{\taum}{\boldsymbol\tau^m}
\newcommand{\Rgb}{\mathbf{R}}
\newcommand{\x}{\mathbf{x}}
\newcommand{\xdot}{\dot{\mathbf{x}}}
\newcommand{\xddg}{\ddot{\mathbf{x}}}

\newcommand{\ct}{\mathbf{c}_t}
\newcommand{\om}{\boldsymbol\omega}

\newcommand{\suprho}{^{\delta\rho} \hspace{0pt}}

\begin{document}

\title{\textbf{Unscented External Force and Torque Estimation for Quadrotors}}


\author{Christopher D. McKinnon and Angela P.~Schoellig
\thanks{The authors are with the Dynamic Systems Lab ({www.dynsyslab.org}) at the University of Toronto Institute for Aerospace Studies (UTIAS), Canada. Email: {chris.mckinnon@mail.utoronto.ca, schoellig@utias.utoronto.ca}}%
\thanks{This work was supported by the Natural Sciences and Engineering Research Council of Canada under the grant RGPIN-2014-04634.}%
\thanks{Video available at: \url{http://tiny.cc/UAV-ForceEstimation}} }

\maketitle

\begin{abstract}
In this paper, we describe an algorithm, based on the well-known Unscented Quaternion Estimator, to estimate external forces and torques acting on a quadrotor. This formulation uses a non-linear model for the quadrotor dynamics, naturally incorporates process and measurement noise, requires only a few parameters to be tuned manually, and uses singularity-free unit quaternions to represent attitude. We demonstrate in simulation that the proposed algorithm can outperform existing methods. We then highlight how our approach can be used to generate force and torque profiles from experimental data, and how this information can later be used for controller design. Finally, we show how the resulting controllers enable a quadrotor to stay in the wind field of a moving fan.
\end{abstract}

%
\IEEEpeerreviewmaketitle

\section{Introduction}
\label{sec:Introduction}

Quadrotors are small and agile, and are becoming increasingly capable for their compact size. They are expected to perform in a wide variety of tasks, where they are either required to physically interact with the environment for applications such as inspection and manipulation \cite{Darivianakis2014,Marconi2012control,YukselReshaping2014,Albers2010semi,Nguygen2013ToolForce}, or fly in close proximity to other quadrotors for applications involving formation flight \cite{Michael2010,Sydney2013,Yeo2015}.  In all these cases, quadrotors may experience significant external forces, which are difficult to model but affect the quadrotor's dynamic behaviour. Accurately estimating external forces and reacting to them appropriately can be essential for completing a given task safely and effectively. The diverse range of applications for quadrotors motivates the development of a method that does not require specialized knowledge about the quadrotor's tasks and detailed dynamic models of the external force effects, which are usually difficult to derive from first principles. This has led to an increased interest in external wrench (that is, combined force and torque) estimators and their application to quadrotors \cite{Nguygen2013ToolForce, TomicCollisionReflexes2014, yuksel2014Force, Augugliaro2013}. 

To date, the most popular approach that incorporates the non-linear dynamics of the quadrotor is to use a non-linear observer. The observer can be applied in conjunction with a feedback controller to improve flight performance in the presence of external forces and torques. While the non-linear observer provides an effective means to estimate forces and torques when they are large and the noise level is low, we found that the accuracy of the estimate degrades quickly as we increased the level of noise. Recovering good performance from the noisy signal required extensive filtering of the inputs and outputs, and tuning the many filter parameters is a time-consuming and intricate process, for which there is no systematic solution.

The goal of this paper is to present an external force and torque estimation algorithm that can provide accurate estimates of the external forces and torques in realistic experimental settings and is parametrized by a small number of effective tuning parameters.
We develop a novel force estimation algorithm borrowing ideas from the Unscented Quaternion Estimator, see \cite{crassidis2003unscented}. We use the estimator to map the forces and torques a quadrotor experiences as it flies close to objects such as a fan. We quantify the accuracy of our estimate and demonstrate in experiment how measurements of this force field can be used in combination with admittance control to track a fan using only the estimated aerodynamic forces.

The first external force estimator for quadrotors was presented in \cite{Augugliaro2013}, focusing on  human-quadrotor interaction. The authors used a Kalman filter and the linearized quadrotor dynamics to estimate external forces, which were then used as input to an admittance controller.  They estimated the full 3-D external force vector, but their analysis was restricted to small changes in attitude due to the linearization.

Since then, researchers have shown numerous applications for quadrotors that use external force and torque estimates as an input to their algorithms and, accordingly, have developed improved force and torque estimators better suited for the demands of these tasks. A non-linear observer was first applied to the task of estimating external forces and torques by \cite{YukselReshaping2014}, and later extended in \cite{TomicCollisionReflexes2014} and \cite{ruggiero2014impedance}. These papers highlight many potential applications including  reducing the risk of damage in a collision, tactile mapping, takeoff and landing detection, identifying the material of a surface by colliding with it \cite{TomicCollisionReflexes2014}, and impedance control \cite{TomicCollisionReflexes2014,ruggiero2014impedance}. Non-linear observers work well in practice if forces are large and sensor noise is small. Otherwise, inputs and outputs of the non-linear observer must be carefully filtered, since the algorithm is based on a deterministic formulation and does not account for process and sensor noise. The filter tuning can be an intricate and time-consuming process. 

Non-linear stochastic state estimation algorithms such as the Unscented Kalman Filter (UKF) are designed to properly handle sensor and process noise, and many of the tuning parameters are derived directly from the noise properties of the sensors making it easy to tune. We demonstrate the effectiveness of such an approach in several experiments with a quadrotor, and also show in simulation a comparison of this algorithm and a non-linear observer.


The first contribution of this work is to design a force estimator that can adequately handle noisy measurements. We develop an external force and torque estimation algorithm based on an Unscented Kalman Filter (UKF) which~\emph{(i)}~uses a non-linear model for the quadrotor dynamics,
\emph{(ii)} explicitly takes into account sensor noise and imperfections in our quadrotor model,
 \emph{(iii)} is light-weight enough to be implemented in applications that require high update rates, and~\emph{(iv)}~uses singularity-free quaternions to represent the attitude of the quadrotor.

The second contribution is to demonstrate in experiment how the estimated values of force and torque can be used to react to a wide variety of aerodynamic disturbances without explicitly modeling them. We include an illustrative experiment where we use force and torque estimates in an admittance controller to enable the quadrotor to track the center of a fan.


 
\section{Force and Torque Estimation}
\label{sec:UKFquaternion}

The first component of the force and torque estimation algorithm is a simple model of the quadorotor dynamics.  External forces and torques are quantities that cannot be explained by our first-principles quadrotor model but are exerted by external sources such as physical contact or air flow induced by a fan.  We present a force/torque estimation scheme based on the Unscented Kalman Filter (UKF) that carefully models the source of process and measurement noise.

The two basic steps in the UKF implementation are the prediction and the correction step. The prediction step predicts the state of the quadrotor at the next time-step given measurements of the motor turn rates and a model of the quadrotor dynamics. In the correction step, the state estimate is updated to better explain the observed vehicle motion. In our case, this means the external force and torque estimates are adjusted to explain differences between measurements of the vehicle position and attitude, and the corresponding predicted values.

\begin{figure}
  \centering
  \includegraphics[width=0.48\textwidth]{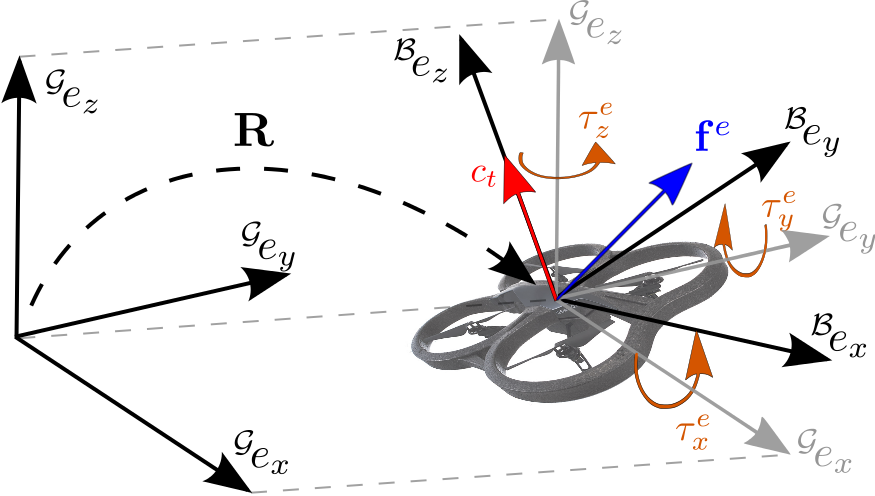}
  \caption[Coordinate frames]{Coordinate frames and variable definitions for the quadrotor with~$\mathcal{B}$ being the body-fixed frame and $\mathcal{G}$ the global frame. The collective thrust $c_t$ produced by the four motors is shown in red. External forces, $\feg = (f^e_x, f^e_y, f^e_z)$, and torques, $\taueg = (\tau^e_x, \tau^e_y, \tau^e_z)$, are shown in blue and orange, respectively, and expressed in global coordinates.}
  \label{fig:CoordFrames}
\end{figure}

\subsection{Prediction Model}

In this section, we present the discrete-time model of the quadrotor dynamics. We use a discrete-time model to accurately represent both the discrete nature of the measurements and inputs, and the corresponding uncertainties in the model. We use subscript $k$ to denote the discrete-time index (i.e., $\mathbf{x}_k = \mathbf{x}(kT)$ with $T$ being the sampling period) and make reasonable assumptions about how quantities vary between time-steps. 


\subsubsection{Translational Dynamics}

The quadrotor is modeled as a rigid body with mass $m$ and $(3\times 3)$ inertia matrix~$\mathbf{I}$. Our model neglects aerodynamic effects, which are reasonably small at slow speeds \cite{Hehn2011}. 

The quadrotor is actuated by four propellers.  Individually, each motor $i$, $i\in\left\{ 1,2,3,4 \right\}$, produces a thrust proportional to the squared motor turn rate \cite{Michael2010}, $c_i = k_i\Omega_i^2$, where the constant $k_i$ may vary depending on the individual propeller efficiency. The individual forces sum up to give the collective thrust,
\begin{align}
  c_t = \sum_{i=1}^{4} k_i \Omega_i^2,
  \label{eqn:MotorThrust}
\end{align}
which acts along the body $z$-axis, $\ebz$, see Fig. \ref{fig:CoordFrames}.

We assume constant acceleration in the global frame between time-steps (that is, constant external force and constant thrust) and neglect the change in direction of $\ct$ in the global frame over one time-step.  These are reasonable assumptions for small time-steps (in our work, $T=5$ ms).  Under these assumptions, the time-discretized translational dynamics are
\begin{align}
  \label{eqn:DiscreteTranslationalDynamics1}
  \x_{k} &= \x_{k-1} + T\xdot_{k-1} + \frac{1}{2} T^2\xddg_{k-1},\\
  \label{eqn:DiscreteTranslationalDynamics2}
  \xdot_{k} &= \xdot_{k-1} + T\xddg_{k-1},\\
  \xddg_{k} &= \Rgb^T_k(\mathbf{c}_{t,k} + \boldsymbol\eta_{c_t,k})/m - \mathbf{g} + \feg_{k}/m,
  \label{eqn:DiscreteTranslationalDynamics}
\end{align}
where $\x_{k} = [x_k,y_k,z_k]^T$ is the position of the center of mass of the quadrotor in global coordinates, $\Rgb^T_k$ is the rotation matrix from the body frame to the global frame, $\feg_k$ is the external force acting on the quadrotor in global coordinates, $\mathbf{g} = [0,0,9.81]^T$ is the gravitational vector, $m$ is the mass of the quadrotor, and $\boldsymbol\eta_{c_t,k}$ is the process noise.

The process noise in \eqref{eqn:DiscreteTranslationalDynamics} is to account for uncertainty in the model of the thrust produced by each propeller, \eqref{eqn:MotorThrust}.  The thrust mapping is derived close to hover and is not accurate when the quadrotor's air speed and attitude are non-zero~\cite{STARMAC2007}.  Moreover, the measurements of the motor turn rates are quantized to 8-bit values which adds quantization noise to the system.  To account for these effects, we add zero-mean Gaussian noise,~$\boldsymbol\eta_{c_t,k}\sim\mathcal{N}(\mathbf{0},\mathbf{Q}_{c_t})$, to the nominal thrust with $\mathbf{Q}_{c_t}$ being the corresponding covariance matrix.  The variance of the first and second element of $\boldsymbol\eta_{c_t,k}$ are non-zero because the estimate of the orientation of $\ebz$ is not perfect, and its orientation with respect to $\mathcal{G}$ changes over one time-step. The third element primarily accounts for uncertainty in the amount of thrust produced by the propellers.

The external forces $\feg_k$ are expressed in the global frame (see Fig. \ref{fig:CoordFrames}). We do not assume any specific underlying dynamics for the external forces. We model their dynamics as a random walk,
\begin{align}
  \feg_{k} &= \feg_{k-1} + \boldsymbol{\eta}_{\feg,k},
  \label{eqn:ForceDynamics}
\end{align}
where $\boldsymbol\eta_{\feg,k}$ is zero-mean Gaussian noise, ${\boldsymbol\eta_{\feg,k}\sim\mathcal{N}(\mathbf{0},\mathbf{Q}_{\feg})}$, and $\mathbf{Q}_{\feg}$ its diagonal covariance matrix.
The expected value of $\feg$ does not change over time but its variance increases. Values farther from the mean become more likely as time passes. This choice for the dynamics of $\feg$ allows the UKF to explain discrepancies between the prediction and measurements by an additional external force acting on the system. The covariance, $\mathbf{Q}_{\feg}$, becomes a tuning parameter. A smaller covariance indicates that we expect the force to change slowly, and a larger covariance means that we expect it to change quickly. A diagonal noise covariance indicates that components of force vary independently. Modeling force dynamics as a random walk has proven sufficient to estimate unknown, changing forces~\cite{yuksel2014Force, Augugliaro2013, TomicCollisionReflexes2014}.

\begin{figure}
  \centering
  \includegraphics[width=0.5\textwidth]{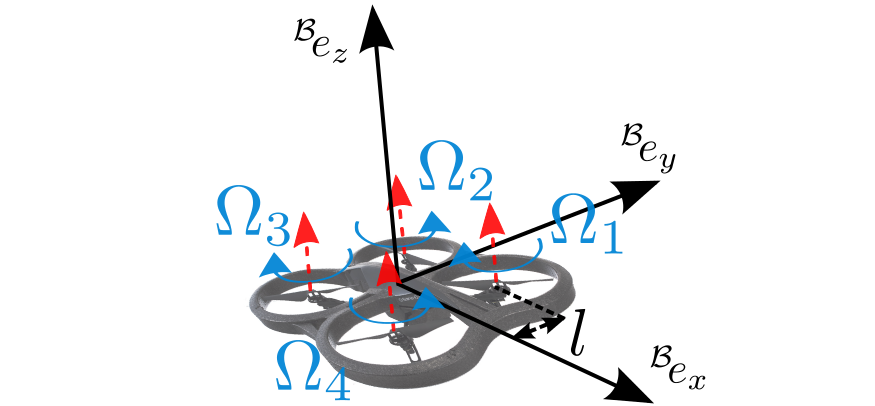}
  \caption[Motor rotation and thrust]{Each motor is a distance $l$ from the $x$- and $y$-axis, and produces a thrust $c_i$ shown in red. The direction of rotation and turn rate, $\Omega_i$, for each motor is shown in light blue.}
  \label{fig:MotorDirections}
\end{figure}

\subsubsection{Rotational Dynamics}
The orientation of the body frame with respect to the global frame can be represented by the $(4 \times 1)$ unit quaternion~$\mathbf{q}~=~[q_0,~\mathbf{q}_v]^T$, 
\begin{equation}
  \mathbf{q} \triangleq \bbm{q_0 \\ \mathbf{q}_v} = \bbm{\cos(\theta/2) \\ \breve{\mathbf{u}}\sin(\theta/2)}.
  \label{eqn:QuaternionDefinition}
\end{equation}
The unit quaternion describes a single rotation by an angle~$\theta$ about an axis~$\breve{\mathbf{u}}$, where the latter is a unit vector expressed in~$\mathcal{G}$. Quaternions have two important properties: they are singularity-free and less susceptible to round-off errors than rotation matrices~\cite{JKelly2011Th}. However, like rotation matrices, unit quaternions must occasionally be re-normalized due to machine precision.  The unit quaternion $\mathbf{q}$ can be converted to the rotation matrix $\mathbf{R}^T$ according to \cite{kellyIJRR2011},
\begin{align}
  \mathbf{R}^T = (2 q_0^2-1)\mathbf{1}_{3 \times 3} + 2 \mathbf{q}_v \mathbf{q}_v^T - 2 q_0 \mathbf{q}_v^{\times},
  \label{eqn:quat2RotMat}
\end{align}
where $\mathbf{q}_v^{\times}$ is the skew-symmetric cross product matrix of $\mathbf{q}_v$,
and $\mathbf{1}_{3\times3}$ is the $(3\times 3 )$ identity matrix.

We model the quadrotor in the `X' configuration, refer to Fig. \ref{fig:CoordFrames} and Fig. \ref{fig:MotorDirections}, where each motor is a distance~$l$ away from the body $x$- and $y$-axis. The motors act in pairs to produce a thrust differential that results in a torque, which is conveniently expressed in the body frame, $\boldsymbol\tau^m~=~[\tau_x^m, \tau_y^m, \tau_z^m]^T$. Referring to \cite{Michael2010} and Fig. \ref{fig:MotorDirections}, the $x$- and $y$-components of $\taum$ are calculated using
\begin{align}
  \tau^m_{x} &= l(k_1\Omega_{1}^2+k_2\Omega_{2}^2-k_3\Omega_{3}^2-k_4\Omega_{4}^2),
  \label{eqn:MotorTauX}\\
  \tau^m_{y} &= l(-k_1\Omega_{1}^2+k_2\Omega_{2}^2+k_3\Omega_{3}^2-k_4\Omega_{4}^2).
  \label{eqn:MotorTauY}
\end{align}
In addition, each motor produces a torque, $M_i, i\in\{1,2,3,4\}$, about its own axis of rotation, which is opposite to its direction of rotation (see Fig. \ref{fig:MotorDirections}). This torque is also proportional to the squared motor turn rate by constants~$p_i$, $ M_i = p_i \Omega_i^2$ \cite{Michael2010}.  Motors 2 and 4 rotate in the positive~$\ebz$ direction opposite to motors 1 and 3. The resulting torque is 
\begin{align}
  \tau^m_{z} = p_1\Omega_{1}^2-p_2\Omega_{2}^2+p_3\Omega_{3}^2-p_4\Omega_{4}^2.
  \label{eqn:MotorTauZ}
\end{align}
The external torque, $\taueg$, which comes from unmodeled external sources, is expressed in the global frame (see Fig.~\ref{fig:CoordFrames}). 

To obtain the discretized rotational dynamics, we assume constant angular velocity in the body frame during each time-step to predict the attitude, and constant motor and external torque during each time-step to predict the angular velocity.  Under these assumptions, the rotational dynamics are
\begin{align}
  \label{eqn:DiscreteRotationalDynamics1}
  \mathbf{q}_{k} &= \boldsymbol\Omega(\om_{k-1}) \mathbf{q}_{k-1}, \\ 
  \om_k   &= \om_{k-1} + T\mathbf{I}^{-1}(\Rgb_{k-1}\taueg_{k-1} + \taum_{k-1} + \boldsymbol\eta_{\taum,k} \nonumber \\
  &\qquad - \om_{k-1}\times \mathbf{I}\om_{k-1}), 
  \label{eqn:DiscreteRotationalDynamics}
\end{align}
where $\om_k = [\omega_{x,k}, \omega_{y,k}, \omega_{z,k}]^T$ is the angular velocity expressed in the body frame, $\mathbf{I}$ is the $(3\times 3)$ inertia matrix, and
\begin{equation}
  \boldsymbol\Omega(\om_k) = 
     \begin{bmatrix} 
       \cos(0.5\norm{\om_k}T) & -\boldsymbol\psi_k^T\\
          \boldsymbol\psi_k   & \cos(0.5\norm{\om_k}T)\mathbf{1}_{3 \times 3} + \boldsymbol\psi_k^{\times}
     \end{bmatrix}
  \label{eqn:applyAngVelQuaternion}
\end{equation}
rotates $\mathbf{q}_{k-1}$ to $\mathbf{q}_{k}$ (see \cite{kellyIJRR2011} equation (5)) with $\boldsymbol\psi_k = \sin(0.5\norm{\om_k})T)\om_k/\norm{\om_k}$, where $\norm{\cdot}$ represents the Euclidean norm \cite{JKelly2011Th}. This is equivalent to multiplying $\mathbf{q}_{k-1}$ by the quaternion rotating through angle $\theta = \norm{\boldsymbol\omega}T$ about axis $\breve{\mathbf{u}}~=~\om_k/\norm{\om_k}$ in the body frame.

The motor torque, $\taum$, is uncertain for the same reasons as $\mathbf{c}_t$ stemming from \eqref{eqn:MotorTauX}-\eqref{eqn:MotorTauZ}, which are derived close to hover.  We model this uncertainty as an additive zero-mean Gaussian noise, $\boldsymbol\eta_{\taum,k} \sim\mathcal{N}(\mathbf{0},\mathbf{Q}_{\taum})$, with $(3 \times 3)$ diagonal covariance matrix, $\mathbf{Q}_{\taum}$.

Similar to the external force, $\feg$, we include an external torque, $\taueg$, and model it as a random walk, where $\boldsymbol\eta_{\taueg,k}$ is zero-mean Gaussian noise, $\boldsymbol\eta_{\taueg,k} \sim\mathcal{N}(\mathbf{0},\mathbf{Q}_{\taueg})$, with $\mathbf{Q}_{\taueg}$ being the diagonal covariance matrix, 
\begin{align}
  \taueg_{k} &= \taueg_{k-1} + \boldsymbol{\eta}_{\taueg,k}.
  \label{eqn:TorqueDynamics}
\end{align}

\subsection{Observation Model}

Measurements come from a high-precision, external, camera-based motion capture system, which measures the full 6-degree-of-freedom pose of the vehicle, $\mathbf{y}_k = (\mathbf{x}_k,\mathbf{q}_k)$, at $200$~Hz. We include additive, zero-mean Gaussian measurement noise for $\mathbf{x}_k$, $\boldsymbol\eta_{\mathbf{x},k} \sim \mathcal{N}(\mathbf{0},\mathbf{G}_{\mathbf{x}})$ and $\mathbf{q}_k$, $\boldsymbol\eta_{\mathbf{q},k}~\sim\mathcal{N}(\mathbf{0},\mathbf{G}_{\mathbf{q}})$. The $(3 \times 3)$ diagonal covariance matrices $\mathbf{G}_{\mathbf{x}}$ and $\mathbf{G}_{\mathbf{q}}$ depend on properties of the camera system. The UKF can be easily extended to include other measurements such as those from a GPS and IMU configuration \cite{simon2006optimal, sinopoli2004kalman}.

\subsection{Unscented Filtering} 
\label{sec:UKF_Algorithm}

The goal of the UKF is to estimate the full state of the system, ${\mathbf{s}}_k = (\mathbf{q}_k,\boldsymbol\omega_k,\mathbf{x}_k,\dot{\mathbf{x}}_k,\boldsymbol\tau^e_k,\mathbf{f}^e_k)$, at each time-step. In this work, we are particularly interested in estimating the external force and torque.  We use an Unscented Kalman Filter approach for reasons highlighted in the introduction.  We choose the UKF over the Extended Kalman Filter (EKF), a common alternative, because of its superior performance on many non-linear problems \cite{JKelly2011Th,batz2013ContactForce}.  The UKF produces an approximation that is accurate to third order for Gaussian random variables, while the EKF is only accurate to first order~\cite{JKelly2011Th}. In addition, the UKF does not require the derivation of analytic Jacobians of the dynamics with respect to the state and process noise, which can be a time-consuming and tedious task for high-dimensional state variables.

The UKF is a recursive Gaussian filter. At each time-step, the probability density function of the state is entirely defined by a mean and a covariance.  The goal at each time-step is to go from a prior belief of the mean and covariance of the state, $\{\hat{\mathbf{s}}_{k-1},\hat{\mathbf{P}}_{k-1}\}$, to a predicted belief, $\{\check{\mathbf{s}}_k,\check{\mathbf{P}}_k\}$, and then correct the prediction using measurements to get the estimate, $\{\hat{\mathbf{s}}_{k},\hat{\mathbf{P}}_{k}\}$, for time-step $k$. We denote  predicted values by $\check{(\cdot)}$ and corrected values by $\hat{(\cdot)}$. The corrected value for one time-step is the prior value for the next.

The UKF uses a special set of points called sigma points to represent uncertainty. These points can be transformed exactly through a non-linearity (e.g., the process or observation model) and then recombined into a mean and covariance to recover a Gaussian probability distribution. Special care must be taken to ensure that uncertainty in the rotational states is properly accounted for during these steps because the Unscented Transform does not account for the unit-norm constraint on quaternions. For this purpose, we follow an approach first presented in \cite{crassidis2003unscented} for spacecraft attitude estimation called the Unscented Quaternion Estimator (USQUE).  
The USQUE represents the mean of a rotational state using singularity-free, unit quaternions with rotational uncertainty represented as a perturbation to the mean parametrized by a $(3 \times 1)$ vector of Modified Rodrigues Parameters (MRPs). MRPs are singular at $\pm 2\pi$ but do not have any constraints and, therefore, may be passed through the Unscented Transform directly \cite{JKelly2011Th}. Uncertainty greater than $\pm 2\pi$ would mean we basically have no knowledge of the attitude of the system, which is usually never the case.

\subsubsection{Preliminaries}
The USQUE requires us to frequently convert between local error quaternions and MRPs. The local error quaternion, 
\begin{equation}
  \delta\mathbf{q} = [\delta q_0, \delta\mathbf{q}_v^T]^T,
  \label{eqn:ErrorQuaternion}
\end{equation}
is used to express a perturbation from the mean attitude estimate.  An error quaternion is converted to an MRP,
\begin{equation}
  \delta \boldsymbol\rho= \frac{\delta \mathbf{q}_v}{1+\delta q_0},
  \label{eqn:mrpError}
\end{equation}
to perform operations involving the Unscented Transform.
An MRP may be transformed back to an error quaternion using\vspace{-6pt}
\begin{equation}
  \delta q_0 = \frac{1-\delta\boldsymbol\rho^T\delta\boldsymbol\rho}{1+\delta\boldsymbol\rho^T\delta\boldsymbol\rho},
  \qquad \delta\mathbf{q}_v =  \delta\boldsymbol\rho(1+\delta q_0),
  \label{eqn:MRP2Quat}
\end{equation}
and can then be added back to the mean rotation.



\subsubsection{Prediction Step}

The first step in each iteration of the UKF is to propagate the prior state estimate to the next time-step using the motion model with noise values set to zero, \eqref{eqn:DiscreteTranslationalDynamics1}-\eqref{eqn:ForceDynamics}, \eqref{eqn:DiscreteRotationalDynamics1}-\eqref{eqn:TorqueDynamics}, and the known input, $\mathbf{c}_{t,k}$ and $\boldsymbol\tau^m_k$, computed from the known turn rates $\Omega_i$ and \eqref{eqn:MotorThrust}, \eqref{eqn:MotorTauX}-\eqref{eqn:MotorTauZ}.
The mean estimate of the state of the system at time-step $k$ is denoted by $ \hat{\mathbf{s}}_k~=~(\hat{\mathbf{q}}_k, \hat{\om}_k,\hat{\x}_k,\hat{\xdot}_k,\hat{\boldsymbol\tau}^e_k, \hat{\mathbf{f}}^e_k)$. The prediction step to go from $\hat{\mathbf{s}}_{k-1}$ to the predicted belief at time $k$, $\check{\mathbf{s}}_{k}$, is outlined below.

The mean prior state $\hat{\mathbf{s}}_{k-1}$ is converted to the minimal $(18 \times 1)$ representation $\suprho\hat{\mathbf{s}}_{k-1} = (\delta\hat{\boldsymbol\rho}_{k-1}, \hat{\boldsymbol\omega}_{k-1}, \hat{\x}_{k-1}, \hat{\xdot}_{k-1}, \hat{\boldsymbol\tau}^{e}_{k-1}, \hat{\mathbf{f}}^e_{k-1})$, where $\delta\hat{\boldsymbol\rho}$ is the $(3 \times 1)$ MRP vector with $\delta\hat{\boldsymbol\rho} = \mathbf{0}$. The MRP vector represents a perturbation from the mean, which is zero for the mean itself.  The state vector is combined with the process noise to form a $(30 \times 1)$ extended state, $\hat{\mathbf{z}}_{k-1} = (\suprho\hat{\mathbf{s}}_{k-1}, \hat{\boldsymbol\eta}_{\tau^m}, \hat{\boldsymbol\eta}_{\tau^e}, \hat{\boldsymbol\eta}_{c_t}, \hat{\boldsymbol\eta}_{\mathbf{f}^e}) = (\suprho\hat{\mathbf{s}}_{k-1}, \hat{\boldsymbol\eta})$, where the process noise has the same mean and covariance for all time-steps. This vector contains all uncertain quantities, where we assume that we know the physical parameters of the system (such as mass and inertia) exactly. However, in general, these may also be included in the estimated state, cf. \cite{QuadParamEstimation2011}. With $\mathbf{\hat{P}}_{k-1}$, the $(18 \times 18)$ covariance matrix for the uncertainty in the prior state $\suprho\hat{\mathbf{s}}_{k-1}$, and $\mathbf{Q}$, the $(12 \times 12)$ stacked process noise covariance constant for all time-steps, the extended mean, $\hat{\mathbf{z}}_{k-1}$, and covariance, $\hat{\boldsymbol\Sigma}_{zz,k-1}$, become
\begin{align}
  \hat{\mathbf{z}}_{k-1} = 
      \begin{bmatrix}
        \hat{\mathbf{s}}_{k-1}\\ \mathbf{0}_{12\times 1}
      \end{bmatrix}, &&
  \hat{\boldsymbol\Sigma}_{zz,k-1} = 
      \begin{bmatrix}
        \mathbf{\hat{P}}_{k-1} & \mathbf{0}_{18\times 12}\\
        \mathbf{0}_{12\times 18} & \mathbf{Q}
      \end{bmatrix}.
\end{align}

With $L=30$ being the dimension of $\hat{\mathbf{z}}_{k-1}$, we compute a set of $(2L+1)$ sigma points, $\mathcal{Z}_{k-1,i}$, $i \in \{1,...,2L+1\}$, according to
\begin{align}
  \label{eqn:GenSigmaPoints1}
  \mathbf{S}_{k-1} \mathbf{S}_{k-1}^T =& \hat{\boldsymbol\Sigma}_{zz,k-1} \\
  \mathcal{Z}_{k-1,0}     =& \hat{\mathbf{z}}_{k-1}\\
  \mathcal{Z}_{k-1,j}    =& \hat{\mathbf{z}}_{k-1} + \sqrt{L+\kappa}\:  col_j\mathbf{S}_{k-1}\\
  \mathcal{Z}_{k-1,j+L}  =& \hat{\mathbf{z}}_{k-1} - \sqrt{L+\kappa}\: col_j\mathbf{S}_{k-1}, \text{$j= 1,\dots, L$},
  \label{eqn:GenSigmaPoints}
\end{align}
where $col_j\mathbf{S}_{k-1}$ is the $j^{th}$ column of the lower triangular matrix from the Cholesky decomposition of $\hat{\boldsymbol\Sigma}_{zz,k-1}$, and $\kappa$ is a tuning parameter, which should be set to two assuming the state follows a Gaussian distribution \cite{StateExtimationLectures}.

Each sigma point is un-stacked into the prior uncertainty and process noise,
\begin{equation}
  \mathcal{Z}_{k-1,i} = 
    \begin{bmatrix}
      \hat{\mathbf{s}}_{k-1,i} \\
      \hat{\boldsymbol\eta}_{k-1,i}
    \end{bmatrix}.
  \label{eqn:UnstackPredSP}
\end{equation}
The MRP vector in each sigma point $i$ is converted to an error quaternion, $\delta \hat{\mathbf{q}}_{k-1,i}$, which is multiplied by the prior mean, $\hat{\mathbf{q}}_{k-1}$, to get the full orientation quaternion for that sigma point,
\begin{equation}
  \hat{\mathbf{q}}_{k-1,i} = \delta\hat{\mathbf{q}}_{k-1,i}\otimes\hat{\mathbf{q}}_{k-1}, \qquad i = 0,\dots,2L,
  \label{eqn:ProcessNoiseUpdate}
\end{equation}
where $\otimes$ represents the quaternion multiplication and adds the rotation $\delta \hat{\mathbf{q}}_{k-1,i}$ to $\hat{\mathbf{q}}_{k-1}$.
This quaternion along with the rest of the states from sigma point $i$ are then passed through the non-linear process model, \eqref{eqn:DiscreteTranslationalDynamics} and \eqref{eqn:DiscreteRotationalDynamics}, to get the predicted state for each sigma point, $\check{\mathbf{s}}_{k,i}$.

Once propagated through the process model, each quaternion is then converted back into an error quaternion, $\delta\mathbf{\check{q}}_{k,i}$, by comparing it to the predicted mean, $\mathbf{\check{q}}_{k,0}$, using
\begin{align}
  \delta\mathbf{\check{q}}_{k,i} &= \mathbf{\check{q}}_{k,i}\otimes[\mathbf{\check{q}}_{k,0}]^{-1},
  \label{eqn:SolveForPropErrorQuat}
\end{align}
and then to MRPs, $\delta\check{\boldsymbol\rho}_{k,i}$, such that each sigma point is now of the form $\suprho\check{\mathbf{s}}_{k,i}$. These sigma points are recombined into the predicted mean and covariance using
\begin{align}
  \suprho\mathbf{\check{s}}_k &= \sum_{i=0}^{2L}\alpha_i\: \suprho\check{\mathbf{s}}_{k,i}, \\
         \mathbf{\check{P}}_k &= \sum_{i=0}^{2L}\alpha_i\: \left( \suprho\check{\mathbf{s}}_{k,i} - \suprho\check{\mathbf{s}}_k\right)\left( \suprho\check{\mathbf{s}}_{k,i} - \suprho\check{\mathbf{s}}_k\right)^T,
  \label{eqn:SigmaPoints2MeanCov}
\end{align}
where
\begin{align}
  \alpha_i = 
  \begin{cases}
    \frac{\kappa}{L+\kappa} & \mbox{if } i = 0,\\ 
    \frac{1}{2}\frac{\kappa}{L+\kappa} & \text{otherwise.}
  \end{cases} 
  \label{eqn:SigmaPointWeights}
\end{align}

Finally, the mean perturbation, $\delta\check{\boldsymbol\rho}_k$, is converted to $\delta \check{\mathbf{q}}_k$ added to $\check{\mathbf{q}}_{k,0}$ yielding the predicted mean state, $\check{\mathbf{s}}_k$, for this time-step.


\subsubsection{Correction Step}

The second step of the UKF is to correct our prediction of the state using measurements of the 6-degree-of-freedom pose, $\mathbf{y}_k = (\mathbf{x}^y_k,\mathbf{q}^y_k)$, obtained from the motion capture system. Here, we could apply a standard Kalman filter update because the observation model is linear, but choose to present the full UKF formulation for completeness and generality. 

The generalized Gaussian correction equations are \cite{StateExtimationLectures},
\begin{align}
  \label{eqn:GeneralKalmanGainEquation}
  \mathbf{K}_k &= \check{\boldsymbol\Sigma}_{xy,k}\check{\boldsymbol\Sigma}_{yy,k}^{-1} \\
  \label{eqn:GeneralPredictedCovarianceUpdate}
  \hat{\mathbf{P}}_k &= \check{\mathbf{P}}_k - \mathbf{K}_k\check{\boldsymbol\Sigma}_{xy,k}^T\\
  \Delta\hat{\mathbf{s}}_k &= \mathbf{K}_k(\mathbf{y}_k - \check{\mathbf{y}}_{k}),
  \label{eqn:GaussianCorrectionEqns}
\end{align}
where $\Delta\hat{\mathbf{s}}_k$ is the correction added to the predicted state, $\mathbf{K}_k$ is the Kalman gain, $\check{\boldsymbol\Sigma}_{xy,k}$ is the predicted state-measurement covariance matrix, and $\check{\boldsymbol\Sigma}_{yy}$ is the predicted measurement covariance matrix to be defined below.

The first step to set up the correction is to form an extended measurement state, $\check{\mathbf{z}}^y_{k} = (\suprho\check{\mathbf{s}}_{k}, \boldsymbol\eta_{x}, \boldsymbol\eta_{\rho})$, which includes the predicted measurement noise. The extended measurement is the predicted mean stacked with a $(6 \times 1)$ vector of zeros representing the mean noise. The extended measurement covariance is a block-diagonal matrix including the predicted uncertainty covariance, $\check{\mathbf{P}}_k$, and the block-diagonal measurement noise covariance, $\mathbf{G}$, containing $\mathbf{G}_{\x}$ and $\mathbf{G}_{\boldsymbol\rho}$, which are constant for all time-steps,
\begin{align}
  \check{\mathbf{z}}^y_{k} = 
      \begin{bmatrix}
        \suprho\check{\mathbf{s}}_k\\ \mathbf{0}_{6 \times 1}
      \end{bmatrix}, &&
    \check{\boldsymbol\Sigma}_{zz,k} = 
      \begin{bmatrix}
        \mathbf{\check{P}}_{k} & \mathbf{0}_{18 \times 6}\\
             \mathbf{0}_{6 \times 18} & \mathbf{G}
      \end{bmatrix}.
  \label{eqn:CorrectionSetup}
\end{align}
The mean and covariance from \eqref{eqn:CorrectionSetup} are converted to a sigma point representation using \eqref{eqn:GenSigmaPoints1}-\eqref{eqn:GenSigmaPoints}.  This gives us a set of predicted sigma points, which include the predicted uncertainty and the measurement noise. These sigma points are passed through the observation model to give us the predicted measurements,
\begin{equation}
  \suprho\check{\mathbf{y}}_{k,i} = 
  \begin{bmatrix}
    \check{\mathbf{x}}_{k,i} + \boldsymbol\eta_{x,i}\\
    \delta{\check{\boldsymbol\rho}_{k,i}} + \boldsymbol\eta_{\rho,i}
  \end{bmatrix}.
  \label{eqn:ObservationEquation}
\end{equation}

The sigma points are recombined into a mean predicted measurement, $\suprho\check{\mathbf{y}}_{k}$, and predicted measurement covariance, $\check{\boldsymbol\Sigma}_{yy,k}$, by using \eqref{eqn:SigmaPoints2MeanCov} and substituting $\check{\mathbf{P}}_k$ with $\check{\boldsymbol\Sigma}_{yy,k}$ and $\suprho\check{\mathbf{s}}_{k}$ with $\suprho\check{\mathbf{y}}_{k}$.
The state-measurement covariance $\check{\boldsymbol\Sigma}_{xy,k}$ is then calculated as
\begin{equation}
  \check{\boldsymbol\Sigma}_{xy,k} = \sum_{i=0}^{2L}\alpha_i(\suprho\check{\mathbf{s}}_{k,i} - \suprho\check{\mathbf{s}}_{k})(\suprho\check{\mathbf{y}}_{k,i} - \suprho\hspace{-2pt}\check{\mathbf{y}}_{k})^T,
  \label{eqn:StateMeasCov}
\end{equation}
where $\alpha_{i}$ are from \eqref{eqn:SigmaPointWeights}.


Now that we have the predicted measurements, we compare it to the actual measurements from our motion capture system.
The measured attitude of the vehicle, $\mathbf{q}_k^y$, is compared to the predicted measurement, $\check{\mathbf{q}}_k$, to get a perturbation,
\begin{equation}
  \delta\mathbf{q}_k^y = \mathbf{q}^y_k \otimes \check{\mathbf{q}}_k^{-1},
  \label{eqn:MeasurementPertubation}
\end{equation}
which is converted to MRPs denoted by $\delta\boldsymbol\rho^y_k$ \cite{crassidis2003unscented}. 
The Kalman gain $\mathbf{K}_k$ and corrected uncertainty $\hat{\mathbf{P}}_k$ are computed using \eqref{eqn:GeneralKalmanGainEquation} and \eqref{eqn:GeneralPredictedCovarianceUpdate}. The correction to the predicted estimate is calculated by comparing the predicted measurement to the actual measurement,
\begin{equation}
    \delta {\mathbf{s}}_k^y
  = \mathbf{K}_k
  \left(
  \begin{bmatrix}
    \mathbf{x}_{k}^y\\
    \delta\boldsymbol\rho^y_{k}
  \end{bmatrix}
   - \suprho\hspace{-2pt}\check{\mathbf{y}}_{k}
   \right).
\end{equation}
MRPs from $\delta{\mathbf{s}}_k^y$ are converted to an error quaternion, $\delta{\mathbf{q}}_k^y$, which is used to update the mean of the predicted attitude. This gives us the corrected attitude for this time-step,
\begin{align}
  \hat{\mathbf{q}}_k &= \delta\mathbf{q}_k^y\otimes\check{\mathbf{q}}_k.
  \label{eqn:CorrectedMean}
\end{align}
The other components of the prediction are updated by direct addition, for example, $ \hat{\x}_k = \delta{\x}^y_k + \check{\x}_k$, completing the measurement update.

\section{Experimental Setup and Calibration}
\label{sec:ExperimentalSetup}

Our experimental platform is the Parrot AR.Drone 2.0 running firmware version 2.3.3. We interface with the AR.Drone through ROS, an open-source robot operating system~\cite{ros}. More precisely, we use ROS Hydro installed on a 64-bit 12.04 Ubuntu operating system.  In addition, we used the ROS \textit{ardrone\_autonomy} package \cite{ros}  version 1.3.1. Measurements of position and attitude are received from the camera system at 200\,Hz.  The vehicle parameters such as mass and rotational inertia are given in \cite{pestana2012ar}.
All experiments were conducted with the indoor hull shown in Fig. \ref{fig:CoordFrames}, which protects the vehicle propellers. 

\begin{figure}
  \centering
  \subfloat{\includegraphics[width=0.45\textwidth]{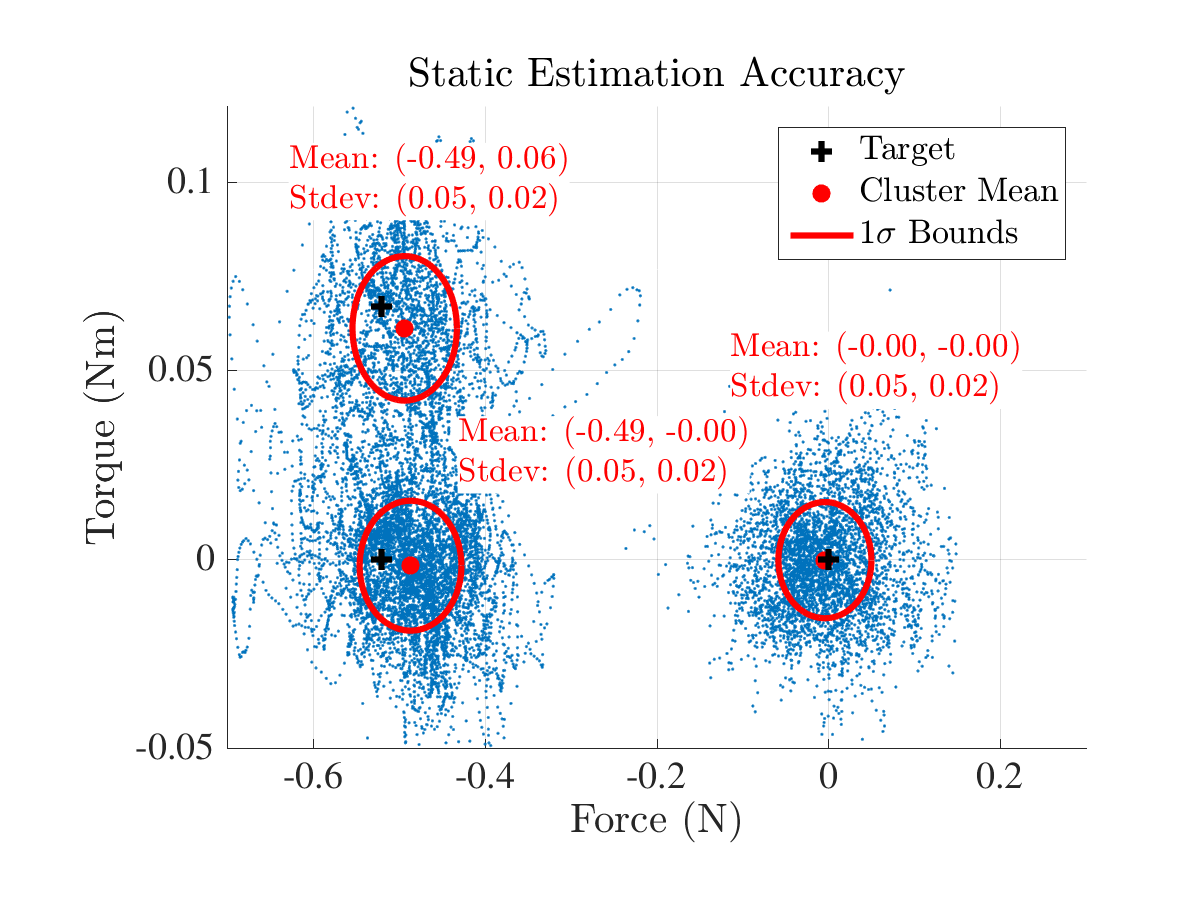}}
  \caption[Steady state accuracy for the force and torque estimate]{Force and torque measurements corresponding to the three test cases. A mixture of Gaussians was fit to the data to get cluster statistics.}
  \label{fig:calibrationGMClusters}
\end{figure}

To quantify the accuracy of our force and torque estimator, we consider three scenarios: hover, suspending a 53 g test mass below the center of mass of the quadrotor, and suspending the test mass under one set of propellers of the quadrotor. The resulting target force-torque values are (0,0), (-0.52,0), and~(-0.52,0.067), respectively. As shown in Fig.~\ref{fig:calibrationGMClusters}, the mean value of the estimates for each scenario is within one standard deviation of the target values, and the standard deviation of the estimates suggests we can measure static force and torque to within 0.05\,N and 0.02\,Nm, respectively.

\begin{figure}
  \centering
  \subfloat{\includegraphics[width=0.45\textwidth]{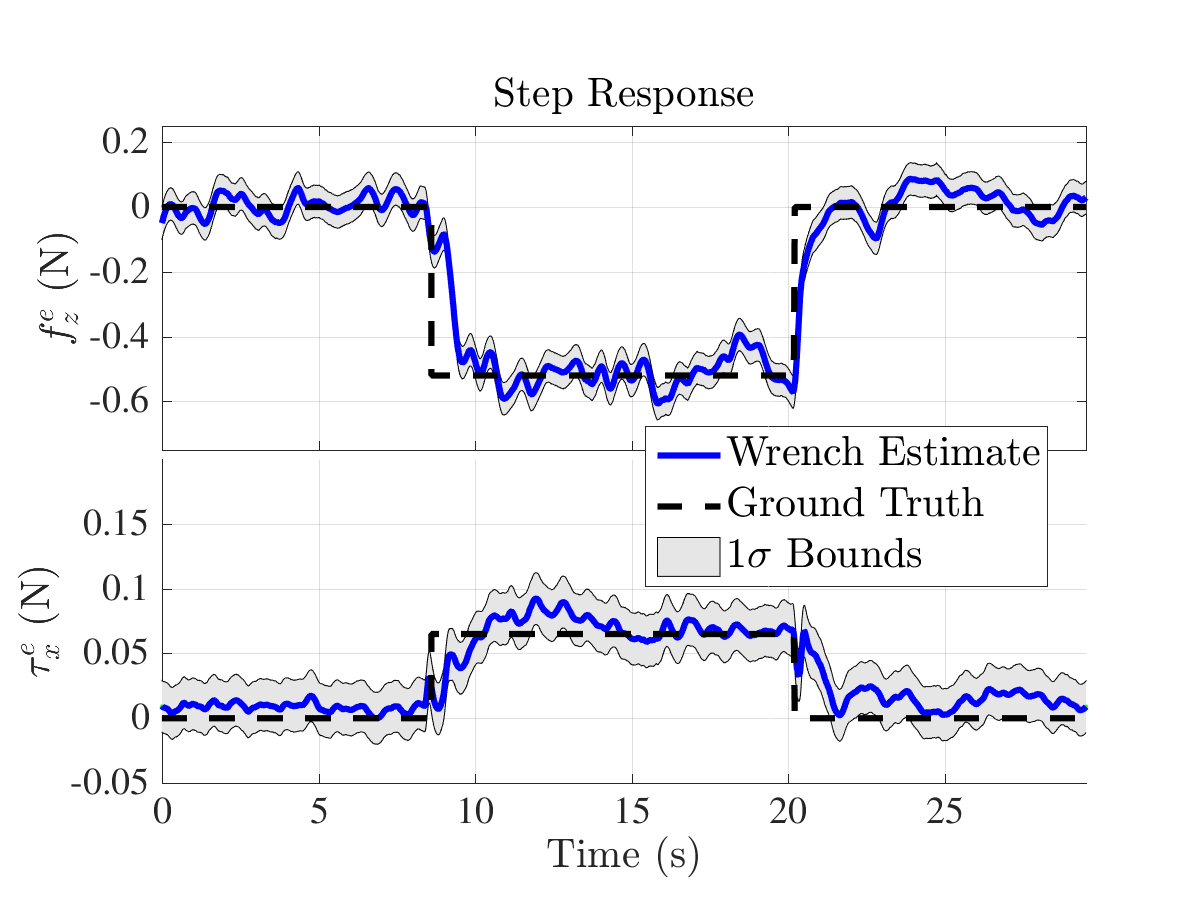}}
\caption[Dynamic response of the force and torque estimates]{Time series showing the step response of the force and torque estimates for a known, applied external force and torque: a mass of 53 g is suspended from the quadrotor at about 7\,s.  The actual external force and torque values are shown in black. In both cases, the estimator converges to the correct value with a rise time of about 1 second.} 
  \label{fig:calibrationTimeSeries}
\end{figure}

We also tested the dynamic response of our estimator, see Fig. \ref{fig:calibrationTimeSeries}. We can achieve a rise time of about 1\,s while retaining good noise suppression characteristics.

\section{Comparison to A Non-Linear Observer}

\begin{figure}
  \centering
  \mbox{
    \subfloat[Force]{\includegraphics[height=0.19\textwidth]{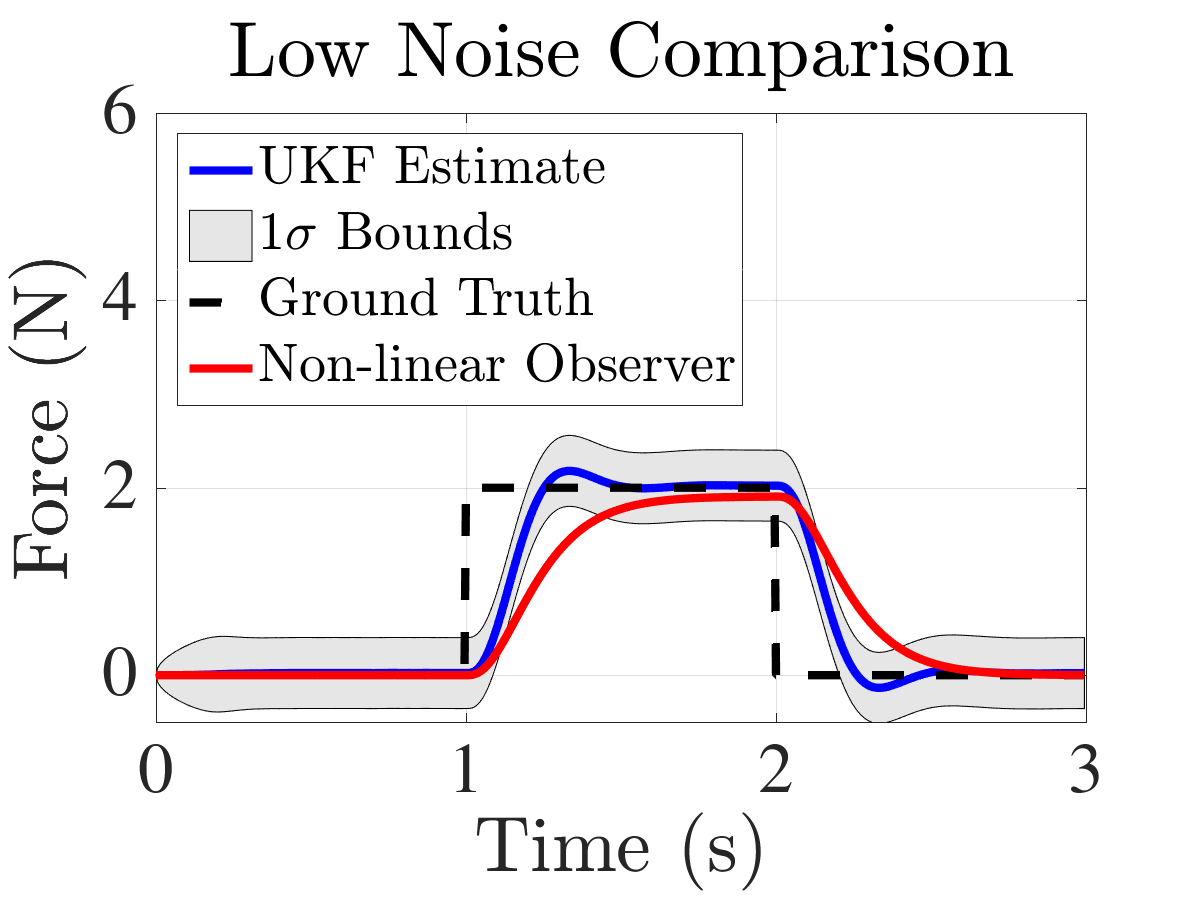}} 
    \subfloat[Torque]{\includegraphics[height=0.19\textwidth]{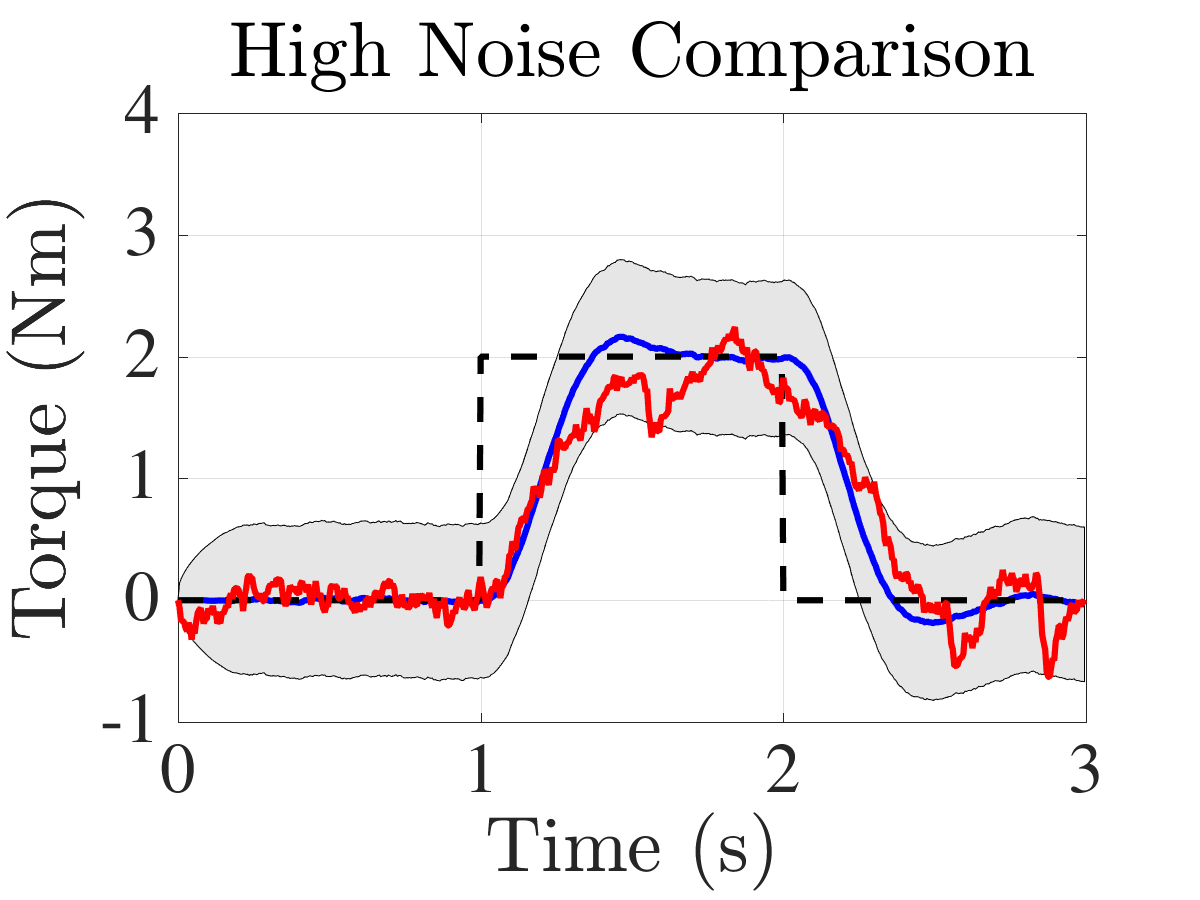}} 
}
\caption[Comparison between the non-linear observer and the UKF-based algorithm]{A direct comparison of the proposed algorithm against a representative non-linear observer. Both algorithms perform well when measurements are relatively noise-free a), however the UKF-based algorithm is more robust to noise as shown in b). For the simulation shown above, zero-mean, Gaussian noise was added to the position and attitude measurements with a standard deviation of 0.01\,m and 0.05\,rad, respectively.}
\label{fig:ForceProfiles}
\end{figure}

We compare our method to the non-linear observer proposed in \cite{yuksel2014Force} with added low-pass filtering on the measurements as suggested in \cite{TomicCollisionReflexes2014}. Filtering was essential for the non-linear observer to produce reasonable results in the presence of noise.  Fig.~\ref{fig:ForceProfiles} depicts simulation results that show how our proposed estimator converges quickly to the true value and remains robust to noise. The non-linear observer can be tuned to perform similar to the proposed estimator when encountering low noise as shown in Fig.~\ref{fig:ForceProfiles}~a). However, the observer is not as effective when the noise increases, see Fig.~\ref{fig:ForceProfiles}~b).


\section{Practical Applications}

Having shown that the force estimator provides reasonable results and having quantified its accuracy, we use it now to measure external forces and torques when we have no specific model for the mechanism causing these forces and torques.  We show how these estimates can be used in the scenario where a quadrotor experiences an aerodynamic disturbance caused by wind from a fan. We encourage readers to check out the associated video at \url{http://tiny.cc/UAV-ForceEstimation}, which includes additional applications.

\begin{figure}
    \mbox{
      \subfloat[Force $f_x^e$]{\includegraphics[height=0.17\textwidth]{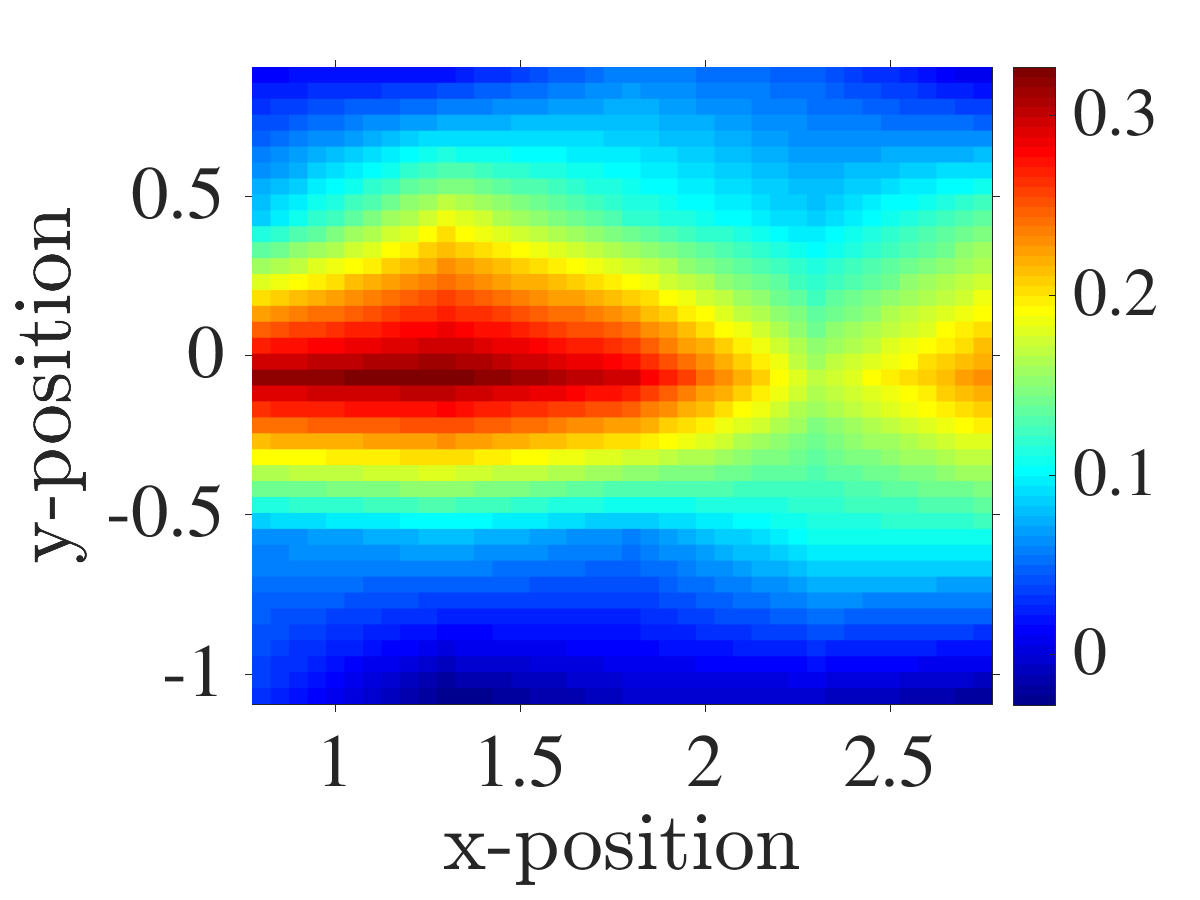}}
      \subfloat[Torque $\tau_z^e$]{\includegraphics[height=0.17\textwidth]{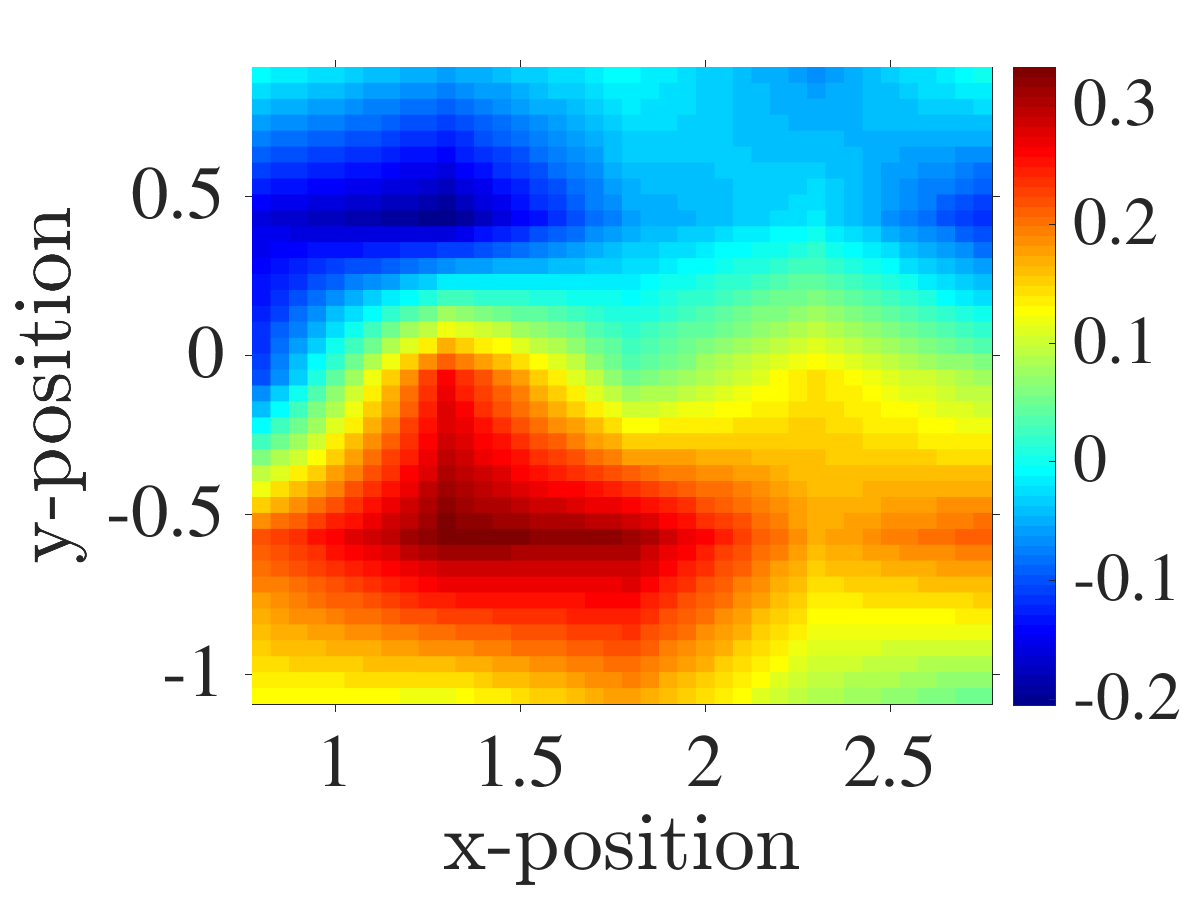}}
    }
  \caption[External force and torque profiles induced by a fan]{Measurements of the force and torque profile in the $(x-y)$ plane using the estimator of Section \ref{sec:UKFquaternion}. There is a strong axial force component in the direction away from the fan, which matches expectations, and an anti-symmetric torque profile. The $\tau_z^e$ profile makes sense intuitively since the fan induces a drag force that, when placed to one side of the center of mass, produces a torque about the body $z$-axis.}
  \label{fig:FanForceProfile}
\end{figure}

We use a fan to generate a large aerodynamic force and show how the torque estimate is reliable enough to guide the quadrotor to the center of the flow using an admittance controller.  Fig. \ref{fig:FanForceProfile} shows two key force and torque profiles due to the fan, estimated using the proposed force estimator. The fan was placed at the origin facing towards the positive $x$-axis, and the quadrotor was flown in a 0.5\,m grid pattern hovering at each point for 5\,s and facing towards the negative $x$-axis. 

We used this visualization of the force and torque induced by the fan to design an admittance controller to track the center of the flow along the $y$-axis. The $\tau_z^e$ profile shown in Fig. \ref{fig:FanForceProfile} is anti-symmetric where the sign of the torque tells us which side of the fan the quadrotor is on. The magnitude of the torque increases to a maximum at roughly 0.5\,m from the axis of the fan.  As a result, we design a proportional, admittance controller, $\dot{y}_{cmd} = k_p \hat{\tau}_z^e$, which can keep the quadrotor in front of the fan by reacting to the torque about the $z$-axis without requiring any specialized wind sensors.

\begin{figure}
  \centering
  \includegraphics[height=0.25\textwidth]{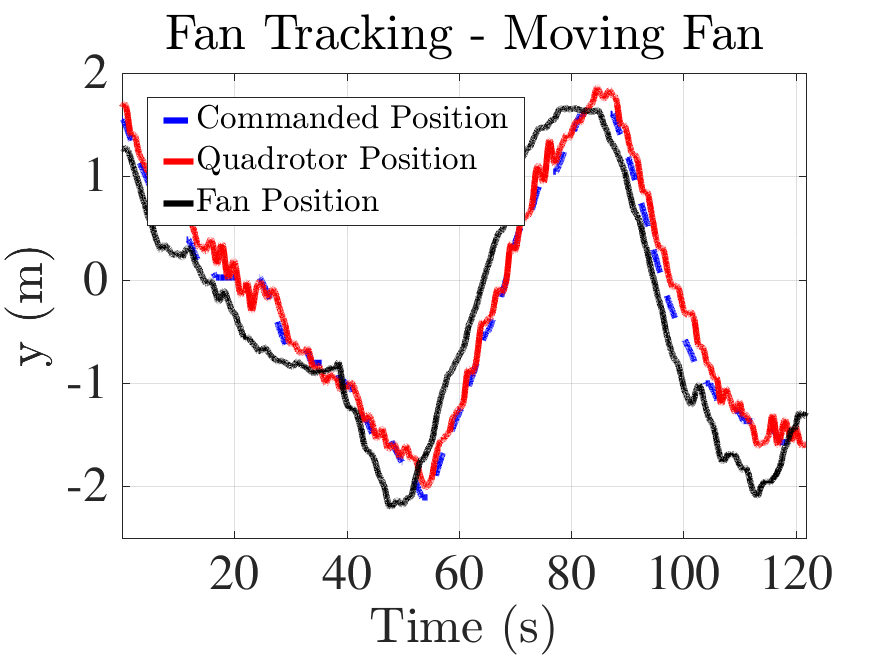}
  \label{fig:DownwashAvoidance}
  \caption[Fan tracking using an admittance controller]{Here we show results of the admittance controller and fan tracking. The fan is 2.3\,m away from the quadrotor moving along the $y$-axis and pointed along the $x$-axis. This shows that the admittance controller can track a moving fan consistently without extra sensors.}
  \label{fig:FanTracking}
\end{figure}

Figure \ref{fig:FanTracking} shows how the force and torque estimator is sensitive enough to allow the quadrotor to move towards the fan from 0.8 m away, and responds quickly enough for the quadrotor to track a moving fan.


\section{Conclusion}

In conclusion, this paper presented an algorithm to estimate external forces and torques acting on a quadrotor. We showed that the proposed algorithm can adequately handle noisy measurements, requires only a few intuitive covariance values to be tuned, and can serve as a basis for reacting to aerodynamic disturbances without relying on specialized knowledge of the underlying dynamics of the disturbance or of the aerodynamic properties of the quadrotor, or on specialized sensors for measuring wind speed. We have demonstrated in experiment how the force estimate may be used in conjunction with an admittance controller to enable a quadrotor to hold position relative to a wind source. We also included a video that shows further applications.

\addcontentsline{toc}{chapter}{Bibliography}
\bibliographystyle{unsrt}
\bibliography{bib/PaperOutline.bib}

\begin{thebibliography}{10}

\bibitem{Darivianakis2014}
G~Darivianakis, K~Alexis, M~Burri, and R~Siegwart.
\newblock {Hybrid Predictive Control for Aerial Robotic Physical Interaction
  towards Inspection Operations}.
\newblock In {\em Proceedings of the International Conference on Robotics and
  Automation}, pages 53 -- 58, 2014.

\bibitem{Marconi2012control}
L~Marconi and R~Naldi.
\newblock {Control of Aerial Robots: Hybrid Force and Position Feedback for a
  Ducted Fan}.
\newblock {\em IEEE Control Systems Magazine}, 32(4):43--65, 2012.

\bibitem{YukselReshaping2014}
B.~Y{\"u}ksel, C.~Secchi, H.~B{\"u}lthoff, and A.~Franchi.
\newblock {Reshaping the Physical Properties of a Quadrotor Through IDA-PBC and
  its Application to Aerial Physical Interaction}.
\newblock In {\em Proceedings of the International Conference on Robotics and
  Automation}, pages 6258--6265, 2014.

\bibitem{Albers2010semi}
A.~Albers, S.~Trautmann, T.~Howard, T.~Hai-Nguyen, M.~Frietsch, and C.~Sauter.
\newblock {Semi-Autonomous Flying Robot for Physical Interaction with
  Environment}.
\newblock In {\em Proceedings of the Conference on Robotics Automation and
  Mechatronics}, pages 441--446, 2010.

\bibitem{Nguygen2013ToolForce}
N.~Hai-Nguyen and L.~Dongjun.
\newblock {Hybrid Force/Motion Control and Internal Dynamics of Quadrotors for
  Tool Operation}.
\newblock In {\em Proceedings of the Intelligent Robots and Systems}, pages
  3458--3464, 2013.

\bibitem{Michael2010}
N.~Michael, D.~Mellinger, Q.~Lindsey, and V.~Kumar.
\newblock {The GRASP Multiple Micro-UAV Testbed}.
\newblock {\em IEEE Robotics \& Automation Magazine}, 17(3):56--65, 2010.

\bibitem{Sydney2013}
N.~Sydney, B.~Smyth, and D.~Paley.
\newblock {Dynamic Control of Autonomous Quadrotor Flight in an Estimated Wind
  Field}.
\newblock In {\em Proceedings of the Conference on Decision and Controls},
  pages 3609 -- 3616, 2013.

\bibitem{Yeo2015}
D.~Yeo, N.~Sydney, and D.~Paley.
\newblock {Onboard Flow Sensing for Downwash Detection and Avoidance with a
  Small Quadrotor Helicopter}.
\newblock In {\em Proceedings of the AIAA Guidance, Navigation and Control
  Conference}, 2015.

\bibitem{TomicCollisionReflexes2014}
T.~Tomic and S.~Haddadin.
\newblock {A Unified Framework for External Wrench Estimation, Interaction
  Control and Collision Reflexes for Flying Robots}.
\newblock In {\em Proceedings of the International Conference on Intelligent
  Robots and Systems}, pages 4197--4204, 2014.

\bibitem{yuksel2014Force}
B.~Y{\"u}ksel, C.~Secchi, H.~B{\"u}lthoff, and A~Franchi.
\newblock {A Nonlinear Force Observer for Quadrotors and Application to
  Physical Interactive Tasks}.
\newblock In {\em Proceedings of the International Conference on Advanced
  Intelligent Mechatronics}, pages 433--440, 2014.

\bibitem{Augugliaro2013}
F.~Augugliaro and R.~D'Andrea.
\newblock {Admittance Control for Physical Human-Quadrocopter Interaction}.
\newblock In {\em Proceedings of the European Control Conference}, pages
  1805--1810, 2013.

\bibitem{crassidis2003unscented}
J.~Crassidis and F.~Landis~Markley.
\newblock {Unscented Filtering for Spacecraft Attitude Estimation}.
\newblock {\em Journal of Guidance, Control, and Dynamics}, 26(4):536--542,
  2003.

\bibitem{ruggiero2014impedance}
F.~Ruggiero, J.~Cacace, H.~Sadeghian, and V.~Lippiello.
\newblock {Impedance Control of VTOL UAVs with a Momentum-Based External
  Generalized Forces Estimator}.
\newblock In {\em Proceedings of the International Conference on Robotics and
  Automation}, pages 2093--2099, 2014.

\bibitem{Hehn2011}
M.~Hehn and R.~D'Andrea.
\newblock {Quadrocopter Trajectory Generation and Control}.
\newblock In {\em Proceedings of the International Federation of Automatic
  Control World Congress}, pages 1485--1491, 2011.

\bibitem{STARMAC2007}
G.~Hoffmann, H.~M.~Huang, S.~L. Waslander, and C.~J. Tomlin.
\newblock {Quadrotor Helicopter Flight Dynamics and Control: Theory and
  Experiment}.
\newblock In {\em Proceedings of the AIAA Guidance, Navigation, and Control
  Conference}, volume~2, 2007.

\bibitem{JKelly2011Th}
J~Kelly.
\newblock {\em {On Temporal and Spatial Calibration for High Accuracy
  Visual-Inertial Motion Estimation}}.
\newblock PhD thesis, University of Southern California, 2011.

\bibitem{kellyIJRR2011}
J.~Kelly and G.~Sukhatme.
\newblock {Visual-Inertial Sensor Fusion: Localization, Mapping and
  Sensor-to-Sensor Self-Calibration}.
\newblock {\em International Journal of Robotics Research}, 30(1):56--79, 2011.

\bibitem{simon2006optimal}
D.~Simon.
\newblock {\em {Optimal State Estimation}}, volume~1.
\newblock John Wiley \& Sons, 2006.

\bibitem{sinopoli2004kalman}
B.~Sinopoli, L.~Schenato, M.~Franceschetti, K.~Poolla, M.~Jordan, and
  S.~Sastry.
\newblock {Kalman Filtering with Intermittent Observations}.
\newblock {\em IEEE Transactions on Automatic Control}, 49(9):1453--1464, 2004.

\bibitem{batz2013ContactForce}
G.~B{\"a}tz, B.~Weber, M.~Scheint, D.~Wollherr, and M.~Buss.
\newblock {Dynamic Contact Force/Torque Observer: Sensor Fusion for Improved
  Interaction Control}.
\newblock {\em International Journal of Robotics Research}, 32(4):446--457,
  2013.

\bibitem{QuadParamEstimation2011}
Q.~Shomin Mellinger, D.~Lindsey and V.~M.~Kumar.
\newblock {Design, Modeling, Estimation and Control for Aerial Grasping and
  Manipulation}.
\newblock In {\em Proceedings of the Intelligent Robots and Systems
  Conference}, pages 2668--2673, 2011.

\bibitem{StateExtimationLectures}
T.~D. Barfoot.
\newblock {\em {State Estimation for Robotics: A Matrix-Lie-Group Approach,
  Available at http://asrl.utias.utoronto.ca/\texttildelow tdb/}}.
\newblock Cambridge University Press, 2016.

\bibitem{ros}
ardrone\_autonomy package, ver. 1.3.1, available at www.ros.org, 2014.

\bibitem{pestana2012ar}
J.~Pestana~Puerta, J.L. S{\'a}nchez~L{\'o}pez, I.~Mellado~Bataller, C.~Fu, and
  P.~Campoy~Cervera.
\newblock {AR.Drone Identification and Navigation Control at CVG-UPM}.
\newblock {\em Journadas Nacionales de Automatica}, 2012.

\end{thebibliography}

\end{document}